\documentclass[letterpaper, 10 pt, conference]{ieeeconf}
\IEEEoverridecommandlockouts
\usepackage{graphicx} 
\usepackage{subcaption} 
\usepackage{xcolor} 
\usepackage{amsmath} 
\usepackage{amssymb}

\usepackage{colortbl}
\usepackage{booktabs}
\usepackage{tabularx}
\usepackage{multirow}
\usepackage{adjustbox}

\makeatletter
\let\NAT@parse\undefined
\makeatother
\usepackage[hidelinks]{hyperref} 
\usepackage[nameinlink,capitalise]{cleveref}
\usepackage{url}

\begin{document}
\title{\LARGE \bf DeFlow: Decoder of Scene Flow Network in Autonomous Driving}

\author{Qingwen~Zhang$^{1}$, Yi~Yang$^{1,2}$, Heng~Fang$^{1}$, 
Ruoyu~Geng$^{3}$, 
Patric~Jensfelt$^{1}$
\thanks{$^{1}$Authors are with the Division of Robotics, Perception, and Learning (RPL), KTH Royal Institute of Technology, Stockholm 114 28, Sweden. (email: qingwen@kth.se)}
\thanks{$^{2}$Authors are with Research and Development, Scania CV AB, Södertälje 151 87, Sweden.}
\thanks{$^{3}$Authors are with Robotics Institute, The Hong Kong University of Science and Technology, Hong Kong SAR, China.}
}
\maketitle

\begin{abstract}
Scene flow estimation determines a scene's 3D motion field, by predicting the motion of points in the scene, especially for aiding tasks in autonomous driving.
Many networks with large-scale point clouds as input use voxelization to create a pseudo-image for real-time running.
However, the voxelization process often results in the loss of point-specific features. 
This gives rise to a challenge in recovering those features for scene flow tasks.
Our paper introduces DeFlow which enables a transition from voxel-based features to point features using Gated Recurrent Unit (GRU) refinement. 
To further enhance scene flow estimation performance, we formulate a novel loss function that accounts for the data imbalance between static and dynamic points. 
Evaluations on the Argoverse 2 scene flow task reveal that DeFlow achieves state-of-the-art results on large-scale point cloud data, demonstrating that our network has better performance and efficiency compared to others. The code is open-sourced at \href{https://github.com/KTH-RPL/deflow}{https://github.com/KTH-RPL/deflow}.
\end{abstract}

\section{Introduction}

Scene flow estimation, which determines the 3D motion field of a scene, is essential in the field of autonomous driving. 
By imitating human behavior when navigating in complex scenes using motion cues, accurate scene flow predictions empower autonomous vehicles (AVs) to interpret and navigate in dynamic environments. 
Such precise estimations further enhance downstream tasks in AVs, encompassing detection, segmentation, tracking, and occupancy flow.

Recent advancements~\cite{luo2021self, najibi2022motion, schmid2023dynablox} highlight the value of class-agnostic motion estimations, which are derived directly from point clouds. 
If satisfactory performance is guaranteed at point-level, the result of scene flow can be effortlessly integrated as a prior for subsequent tasks like prediction and detection~\cite{najibi2022motion}.
This technique will potentially contribute to elevating the efficiency and adaptability of autonomous driving systems in dynamic scenarios.

\begin{figure}[t]
\centering
\includegraphics[trim=280 250 280 0, clip, width=\linewidth]{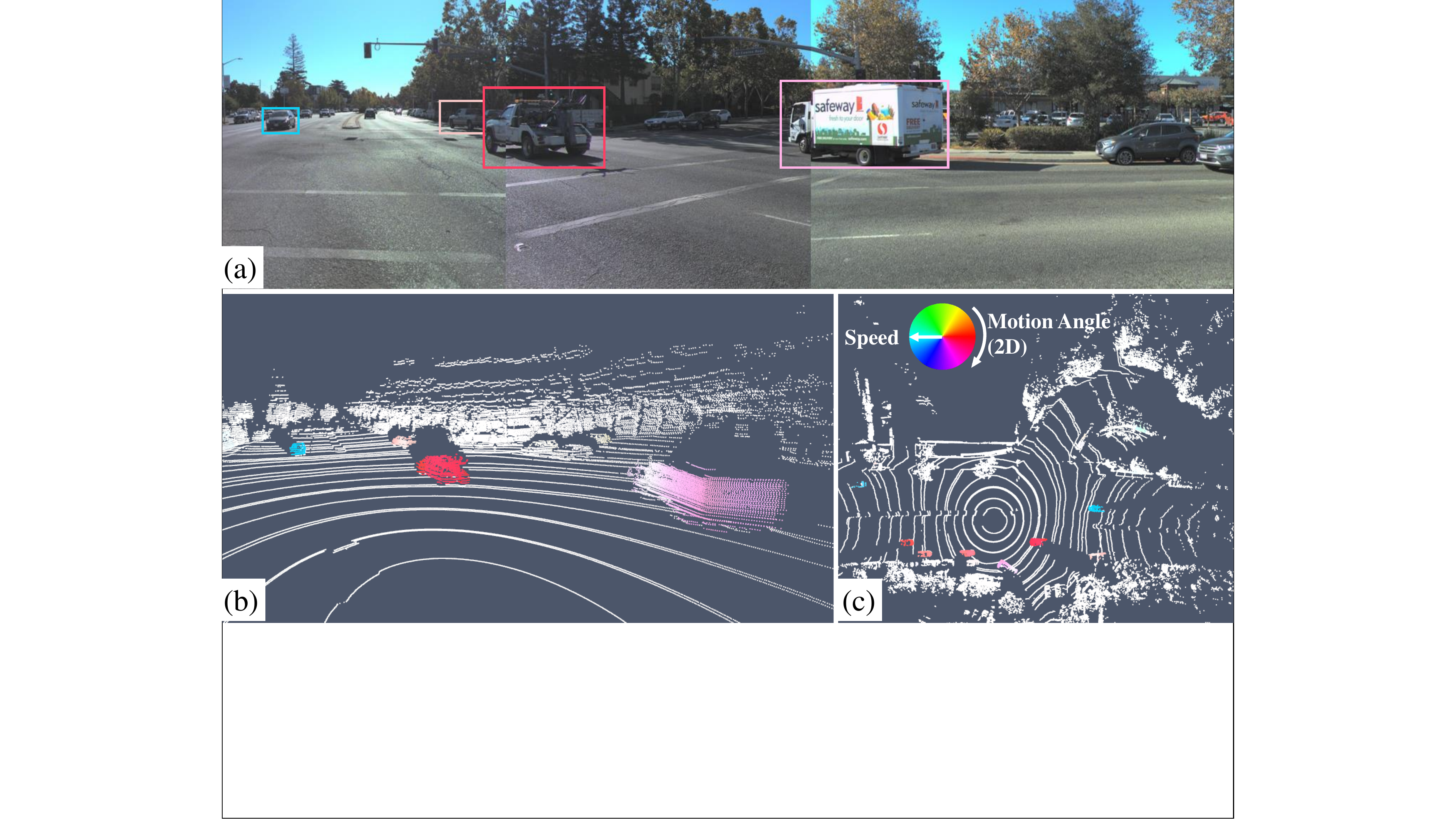}
\caption{LiDAR scene flow estimation using our DeFlow method on the Argoverse 2. The predicted scene flow for each point is color-coded based on direction, with the color wheel anchored in the world frame. (a) Camera view for visualization purposes only. (b)(c) Estimated LiDAR point clouds' flow. Varied colors represent different directions, with more saturated colors indicating higher velocities. (b) Front view. (c) Bird's-eye view.}
\label{fig:backgroud}
\vspace{-1.0em} 
\end{figure}

Most methods~\cite{scoop, wei2020pv, wang2023dpvraft} in object registration scene flow focus on relatively small-scale point cloud data like synthetic datasets Shapenet~\cite{chang2015shapenet} and FlyingThing3D~\cite{mayer2016large}. 
When they use point cloud data in autonomous driving~\cite{fastflow3d, menze2015object}, the points are downsampled to a size of 8,192 points or less.
These methods fail, due to memory overflow on modern driving datasets with the full number of points as input. Datasets like Argoverse2~\cite{Argoverse2} and Waymo~\cite{fastflow3d} are closer to the real autonomous vehicles' sensor setup, where the number of points in one frame is around 80k-177k. 
Recently, some methods~\cite{li2021neural, chodosh2023re} have employed Multi-Layer Perceptrons (MLPs) to optimize proposed self-supervised objective functions that can successfully run on various sizes of datasets. 
However, their runtimes extend from 26 to 35 seconds per frame~\cite{zeroflow}.
In the field of autonomous driving, real-time capability is important. 
Consequently, these optimization-based methods fall short of practicality. 

Given the necessity to process and estimate scene flow on full large point cloud datasets in real-time, FastFlow3D~\cite{fastflow3d} emerges as a practical solution.
An essential strategy to achieve the real-time requirement is voxelization. It is a popular point cloud processing technique, particularly for detection tasks~\cite{qi2017pointnet, zhou2020end, lang2019pointpillars}.
However, there is a distinct difference between detection and scene flow tasks: the latter necessitates point-level results.
Voxelization-based methods often fail to realize the importance of the decoder design in the scene flow task, resulting in their inability to differentiate features among points contained within the same voxel.
This is because all points within a given voxel inherit the same features from the convolutional network.

Addressing these challenges, we present DeFlow which conducts the Gated Recurrent Unit (GRU) refinement module to reconstruct the different features of points inside the same voxel, markedly improving the final results. 
We evaluated our method using the Argoverse 2 scene flow task, and it achieved state-of-the-art results on the online leaderboard, leveraging a training set of 100k labeled frames. An example is shown in~\cref{fig:backgroud}. Our approach is available open-source at \color{blue}\href{https://github.com/KTH-RPL/deflow}{https://github.com/KTH-RPL/deflow}\color{black}. In summary, our primary contributions include:

\begin{itemize}
\item The introduction of a novel real-time network that integrates GRU with iterative refinement in the decoder design, effectively transitioning from voxel to point features.
\item The proposal of a new loss type optimized for imbalanced data distribution on static and dynamic points.
\item Achieving state-of-the-art results on the large-scale point cloud dataset Argoverse 2 online leaderboard.
\end{itemize}

\section{Related work}
\label{sec:related_work}
Introduced by Vedula \textit{et al.}~\cite{vedula2005three}, scene flow estimation captures the 3D point motion field of a scene. This concept evolved from the two-dimensional optical flow estimation, a classical topic that predicts apparent motion patterns in 2D images.

Existing methodologies can be broadly categorized into optimization-based and learning-based approaches. A prominent example within the optimization category is NSFP~\cite{li2021neural}. This method utilizes MLPs to refine the flow, leveraging the chamfer distance as a metric. 
Such methods are typically not categorized as learning-based approaches since they do not save learned weights to infer on other frames. Instead, they employ MLPs to optimize each frame. The considerable inference time of NSFP makes it impractical for real-time applications.

Recent 3D learning-based approach~\cite{wei2020pv, scoop, lu2022gma3d, cheng2022bi} can be traced back to 2D approaches~\cite{teed2020raft, xu2022gmflow, sui2022craft}. 
RAFT~\cite{teed2020raft}, a notable work in optical flow is a good representative. It constructs a multi-scale 4D correlation volume for pixel pairs by iteratively updating the flow field through a recurrent unit. 
Based on it, PV-RAFT~\cite{wei2020pv}, DPV-RAFT~\cite{wang20233d} adapt their frameworks for 3D point cloud data. 
However, these methods become computationally challenging for recent large-scale points due to memory overflow. The issue often happened during the construction of distance matrix~\cite{wei2020pv, wang20233d} or correlation matrix~\cite{scoop} that grow exponentially with the number of points.

\begin{figure}[t]
\centering
\includegraphics[width=\linewidth]{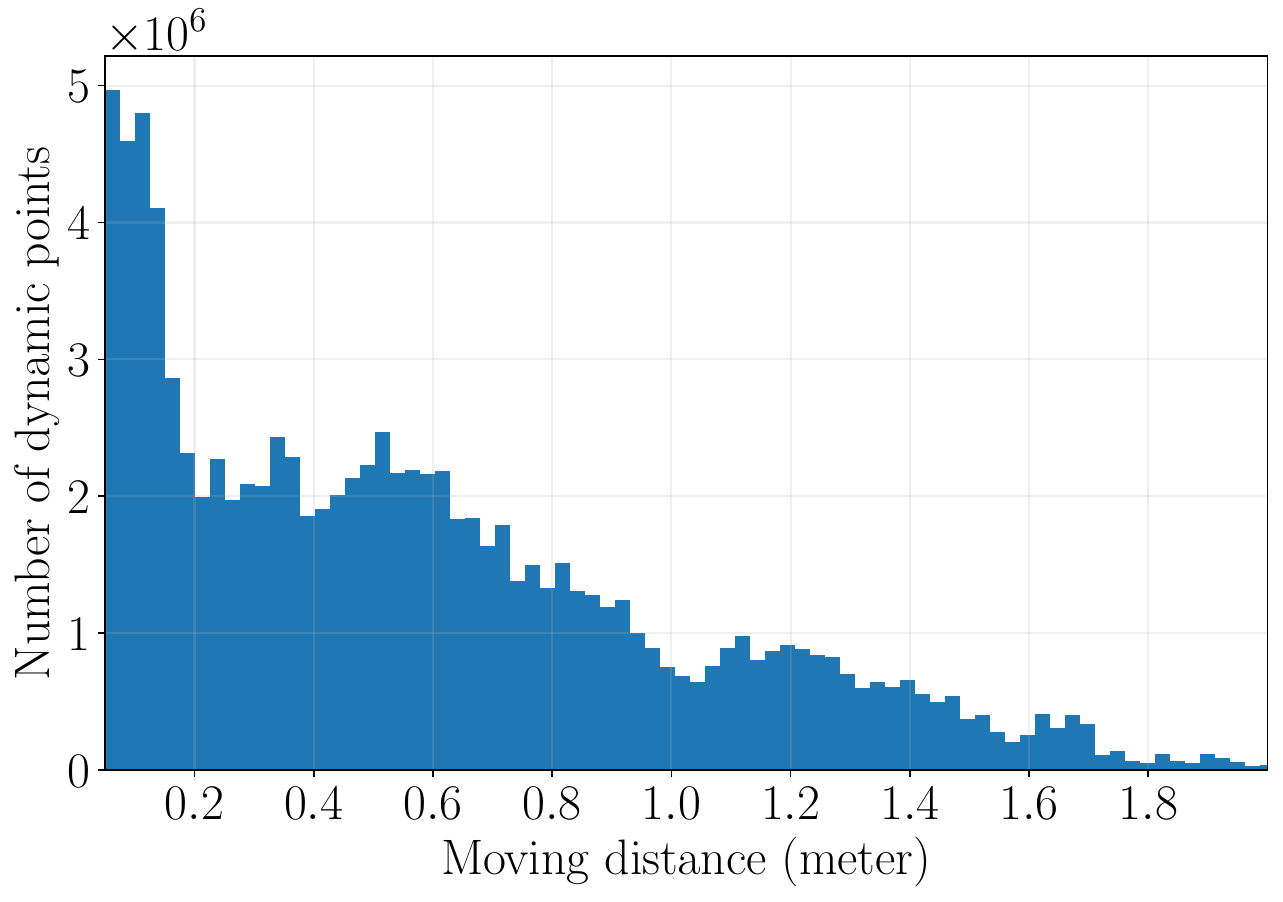}
\caption{
Histogram of moving distances in 0.1$\mathrm{~s}$ for all dynamic points across all scenes in the Argoverse 2 validation dataset (10$\mathrm{~Hz}$).
The x-axis represents the distance in meters, ranging from 0.05 to 2.0 meters. The y-axis indicates the number of points for each distance range. 
The dynamic points are densely distributed within 0.2 meters.
}
\label{fig:dynamic_dis}
\vspace{-1.0em} 
\end{figure}

Fortunately, there are some encoder designs for processing large point clouds in other point cloud related tasks like detection~\cite{qi2017pointnet, zhou2020end, lang2019pointpillars}, and segmentation~\cite{pvkd, openworld2022jun}. 
PointPillar proposed by Lang \textit{et al.}~\cite{lang2019pointpillars}, stands out as one of the most favored encoders due to its impressive performance combined with high efficiency.
The approach, which transforms points to voxels, creating pseudo-images for convolutional networks, is inspired by the 2D method: FlowNet~\cite{dosovitskiy2015flownet}, the pioneering CNN for optical flow estimation, employs a U-Net autoencoder architecture. 
In scene flow task, FastFlow3D~\cite{fastflow3d}, a network that can run on large point cloud data (80k - 177k points) in real-time, also uses PointPillar as its encoder. 
However, the FastFlow3D decoder struggles to differentiate features among points within the same voxel, a limitation attributed to voxelization in PointPillar. 
Using the PointPillar approach can improve efficiency but might reduce performance if the design of the decoder is not thought out well.

It is applicable to adapt the decoder design developed in optical flow to the ones in the scene flow task. 
Previous networks used in optical flow like PWC-Net~\cite{sun2018pwc} employ a context network to expand the receptive field size of outputs, refining flow through dilated convolutions~\cite{yu2015multi}. Moreover,
IRR~\cite{hur2019iterative} offers iterative and residual refinement, achieving enhanced accuracy without enlarging the network. 
Feihu \textit{et al.}~\cite{zhang2021separable} utilizes the Gated Recurrent Unit (GRU) for refinement, iterating multiple times to estimate flow. 
Following the line of the works in optical flow, we adopt GRU in our method to emphasize the transition from voxel to point features in the 3D point scene flow task.

\begin{figure*}[t!]
\centering
\includegraphics[width=\linewidth]{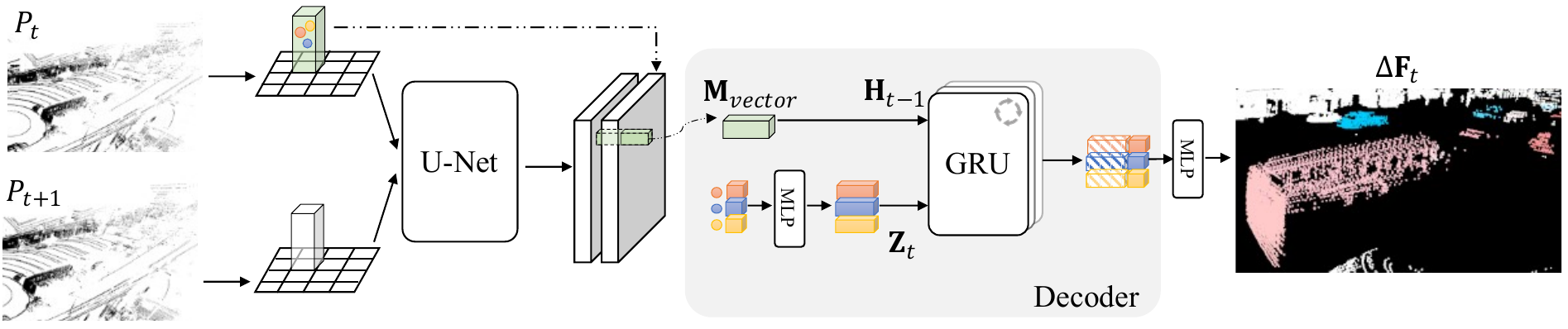}
\caption{\textbf{DeFlow Architecture}. The feature-extracting step, derived from PointPillars, takes two consecutive point clouds as input and transforms them into voxels. The encoder utilizes a convolutional U-Net backbone.
Our novel decoder merges the encoder output with the point offset from PointPillars, employing a GRU for refinement. This process reconstructs the voxel-to-point information, ultimately producing the flow result.
}
\label{fig:arc}
\end{figure*}

\section{Problem Statement}
Our research tackles the challenge of real-time scene flow estimation in autonomous driving. Given two sequential input point clouds, $\mathcal{P}_t$ and $\mathcal{P}_{t+1}$, captured at times $t$ and $t+1$ respectively, along with the ego movement as the transformation matrix $\mathbf{T}_{t,t+1}$, the objective is to predict the motion vector as flow $\hat{\mathbf{F}}_{t,t+1}(p) = (x,y,z)^T$ for each point $p \in \mathcal{P}_t$.

Knowing the frequency of our sensor data collection (10$\mathrm{~Hz}$), it becomes straightforward to interpret the flow as velocity. 
The overarching goal is to minimize the End Point Error (EPE) which presents the discrepancy between the predicted flow and the ground truth flow, as expressed by the following equation:
\begin{equation}
    \min \underbrace{ \frac{1}{\left\|\mathcal{P}_t\right\|} \sum_{p \in \mathcal{P}_t} \left\| \hat{\mathbf{F}}(p) - \mathbf{F}_{gt}(p) \right\|_2 }_{\text{EPE}}.
    \label{eq:opti}
\end{equation}

\section{Approach}
Our methodology builds upon the overall pipeline of FastFlow3D~\cite{fastflow3d}, which was designed specifically for large point cloud data. 
Briefly, it first voxelizes points and formats them into a bird-eye-view grid feature with a specified resolution. Through convolutional layers, the network efficiently learns features within these voxels.

Based on our analysis of Argoverse 2 (as shown in~\cref{fig:dynamic_dis}), we observed that the dynamic points are densely distributed within the range of 0.05 to 0.2$\mathrm{m}$ movement.
Consequently, the choice of resolution is important, as demonstrated in~\cref{sec:quan}.
However, a higher resolution also leads to an increase in computational demands, making it impractical for low-computational devices. 
If a coarse resolution is chosen, the design of the decoder becomes crucial to differentiate point features within the same voxel to achieve results on par with those of a finer resolution.

In the subsequent sections, we outline our framework as depicted in~\cref{fig:arc} and highlight the areas of focus and improvement.
\subsection{Input and Output}
Leveraging HD maps provided by most autonomous driving datasets~\cite{sun2020scalability, Argoverse2}, or by deploying ground segmentation techniques like~\cite{lee2022patchworkpp}, we can easily exclude ground points from both $\mathcal{P}_t$ and $\mathcal{P}_{t+1}$.

The flow $\hat{\mathbf{F}}$ from $\mathcal{P}_t$ to $\mathcal{P}_{t+1}$ is decomposed into two parts as following:
\begin{equation}
    \hat{\mathbf{F}} = \mathbf{F}_{ego} + \Delta \hat{\mathbf{F}}, 
\end{equation}
where $\mathbf{F}_{ego}$ is the flow resulting in ego vehicle's motion which can be directly obtained from $\mathbf{T}_{t,t+1}$, and $\Delta \hat{\mathbf{F}}$ is our network output.

\subsection{Encoder and Backbone}
For point cloud rasterization, we use the dynamic voxelization technique from PointPillars~\cite{zhou2020end, lang2019pointpillars} that improves the framework's efficiency. 
We compute each point's offset from the pillar center and the cluster offset from the point to its cluster coordinates. Post voxelization, a linear transformation aggregates all points within a pillar.

After encoding $\mathcal{P}_t$ and $\mathcal{P}_{t+1}$ into grids, we use a 2D convolutional U-Net backbone. Both grids undergo processing through this shared-weight backbone. 
\begin{table*}[t!]
\centering
\def\arraystretch{1.2}
\begin{tabular}{clccccccc} 
\toprule
\multicolumn{2}{c}{Method}               & \multicolumn{1}{l}{EPE 3-Way ↓} & \multicolumn{1}{l}{EPE FD ↓} & \multicolumn{1}{l}{EPE BS ↓} & \multicolumn{1}{l}{EPE FS ↓} & \multicolumn{1}{l}{\begin{tabular}[c]{@{}l@{}}Dynamic\\IoU ↑\end{tabular}} & \multicolumn{1}{l}{\begin{tabular}[c]{@{}l@{}}Dynamic\\AccRelax ↑\end{tabular}} & \multicolumn{1}{l}{\begin{tabular}[c]{@{}l@{}}Dynamic\\AccStrict ↑\end{tabular}}  \\ 
\hline
\multicolumn{2}{c}{FastNSFP~\cite{li2023fast}}             & 0.1657                        & 0.3540                      & 0.1025                     & 0.0406                     & 0.0924                                                                   & 0.3729                                                                        & 0.1958                                                                          \\ 
\cline{1-2}
\multicolumn{2}{c}{NSFP w Motion Comp~\cite{li2021neural}}   & 0.0685                        & 0.1503                     & 0.0248                     & 0.0302                     & 0.3199                                                                   & \underline{0.6956}                                                                        & \textbf{0.4537}                                                                 \\ 
\cline{1-2}
\multirow{2}{*}{Zeroflow~\cite{zeroflow}} & Standard     & 0.0814                        & 0.2109                     & 0.0080                      & 0.0254                     & 0.4791                                                                   & 0.4363                                                                        & 0.1873                                                                          \\
                          & XL           & \underline{0.0569}                        & \underline{0.1440}                      & 0.0089                     & \textbf{0.0178}                     & 0.5224                                                                   & 0.6106                                                                        & 0.3219                                                                          \\ 
\cline{1-2}
\multicolumn{2}{c}{FastFlow3D~\cite{fastflow3d}}           & 0.0782                        & 0.2073                     & \textbf{0.0020}                     & 0.0253                     & \underline{0.5760}                                                                   & 0.4407                                                                        & 0.1965                                                                          \\
\cline{1-2}
\multicolumn{2}{c}{DeFlow (Ours)}        & \textbf{0.0534}               & \textbf{0.1340}             & \underline{0.0029}                     & \underline{0.0232}                     & \textbf{0.6289}                                                          & \textbf{0.7213}                                                               & \underline{0.4483}                                                                          \\
\bottomrule
\end{tabular}
\caption{
Comparisons on Argoverse 2 sensor test set in the online leaderboard~\cite{onlineleaderboard}. 
Our methods achieve state-of-art performance in the scene flow task. The main improvement happens in the accuracy of flow estimation on dynamic points, with larger improvements in the Dynamic IoU and Dynamic Accuracy Relaxed. We \textbf{bold} the best results and \underline{underline} the second best results.}
\label{tab:main_res}
\vspace{-1.0em} 
\end{table*}
\subsection{Decoder}
The process of obtaining point-wise flow in FastFlow3D is facilitated by the unpillar operation. This operation, for each point, retrieves the associated flow embedding grid cell, appends the point feature, and employs a multi-layer perceptron to deduce the flow vector. However, as previously highlighted, this approach is not appropriately designed for the reconstruction of voxel-to-point features.

As illustrated in~\cref{fig:arc}, prior to the decoder's operation, we concatenate the pillar features of $\mathcal{P}_t$ with the output of the U-Net features, 
resulting in a format of $\mathbf{M}_{vector} \in \mathbb{R}^{N \times C}$. Here, $N$ represents the number of points in $\mathcal{P}_t$, and $C$ denotes the number of feature channels post concatenation. 
We observed that simply concatenating the $C$ channels common to all points in a voxel with the 3 channels dedicated to point offset led to an imbalance.
For the points in the same voxel, the majority of the channels are identical.

An intuitive solution to this imbalance is to expand the point offset features to match the dimensionality of $C$. However, this modification worsen the original performance, proving that a dedicated network design is necessary.

Drawing inspiration from the 2D optical flow techniques that utilize GRU~\cite{zhang2021separable}, we propose an alternative method in 3D scene flow depicted in~\cref{fig:arc} decoder. In this approach, we designate $\mathbf{M}_{vector}$ as the first hidden states, denoted as $\mathbf{H}_0$. By employing a linear layer, we expand the point offsets and set them as input, represented as $\mathbf{x}$. 
The relationship between these components is captured by:
\begin{equation}
    \mathbf{H}_t=\mathbf{Z}_t \odot \mathbf{H}_{t-1}+\left(1-\mathbf{Z}_t\right) \odot \tilde{\mathbf{H}}_t ,
\end{equation}
where $\mathbf{Z}_t$ serves as the update gate. It takes $\mathbf{x}$ as its input, which subsequently undergoes processing via 1D convolution layers, utilizing a Sigmoid activation function.
The term $\mathbf{H}_{t-1}$ represents the previous hidden state. For the initial instance, $\mathbf{H}_0$ is set to $\mathbf{M}_{vector}$.
$\tilde{\mathbf{H}}_t$ stands for the candidate hidden state, which is determined by the reset gate and the model weights.

To maintain a small model without inflating the number of parameters, we employ multiple iterations within GRU layers. 
After completing these iterations, the most recent hidden state $\mathbf{H}_t$ is concatenated with the point offset features. This combined entity then proceeds through MLPs, resulting in the generation of the final delta flow, denoted as $\Delta \hat{\mathbf{F}}$ as shown in~\cref{fig:arc}.
A comprehensive ablation study detailing the intricacies and performance of our decoder design is presented in~\cref{sec:quan}.

\subsection{Loss Function}
\label{sec:loss_fn}

The task of scene flow estimation in autonomous driving scenarios is inherently challenging due to the dynamic nature of the environment. A significant portion of the LiDAR points, which reflect static structures such as buildings or roads, remain stationary. This leads to a label imbalance in the dataset, with more background static points than others. To address this, the loss function incorporates a scaling function, denoted as $\sigma(p)$, to balance the contribution of each point based on its motion characteristics:
\begin{equation}
    \mathcal{L} = \frac{1}{\left\|\mathcal{P}_t\right\|} \sum_{p \in \mathcal{P}_t} \sigma(p)\left\| \Delta \hat{\mathbf{F}}(p)- \Delta \mathbf{F}_{gt}(p)\right\|_2,
\end{equation}
where $\left\|\mathcal{P}_t\right\|$ is the number of points in $\mathcal{P}_t$.

FastFlow3D~\cite{fastflow3d}, in their experiments, introduced a scaling approach based on the distinction between foreground and background points. 
The difference between the two is determined by whether a point is contained within the bounding box of any tracked object.
\begin{equation}
    \sigma(p)_{t}= \begin{cases}1 & \text { if } p \in \text { Foreground } \\ 0.1 & \text { if } p \in \text { Background }\end{cases}
    \label{eq:scaled_type}
\end{equation}

With the advent of self-supervised learning, where labels distinguishing foreground and background are absent, Zeroflow~\cite{zeroflow} proposed an alternative scaling function. This function scales based on the speed (flow) of the point's motion:
\begin{equation}
    \sigma(p)_{s}= \begin{cases}0.1 & \text { if } s(p)<0.4 \mathrm{~m} / \mathrm{s} \\ 1.0 & \text { if } s(p)>1.0 \mathrm{~m} / \mathrm{s} \\ 1.8 s-0.8 & \text { o.w. }\end{cases}
    \label{eq:scaled_speed}
\end{equation}

Building upon the insights from Zeroflow, we propose a nuanced scaling approach that takes into account the distribution of dynamic and static point numbers. This approach divides $\mathcal{P}_{t}$ into three categories based on their motion speed $\{\mathcal{P}_{t/1}, \mathcal{P}_{t/2}, \mathcal{P}_{t/3}\}$, as defined in~\cref{eq:scaled_speed}. 
The total loss is then the sum of the losses from these three categories:
\begin{align}
    \mathcal{L}_{total} = \sum_{i=1}^{3}\frac{1}{\left\|\mathcal{P}_{t/i}\right\|} \sum_{p \in \mathcal{P}_{t/i}} \left\| \Delta \hat{\mathbf{F}}(p)- \Delta \mathbf{F}_{gt}(p)\right\|_2.
    \label{eq:scaled_three}
\end{align}

This comprehensive loss function ensures that the model balances different types of point motion, providing a robust estimation of scene flow in diverse scenarios.
\begin{table}[t]
\centering
\def\arraystretch{1.2}
\begin{tabular}{ccccc} 
\toprule
\multicolumn{1}{l}{\# Itr} & \multicolumn{1}{l}{EPE 3-way ↓} & \multicolumn{1}{l}{EPE FD ↓} & \multicolumn{1}{l}{EPE BS ↓} & \multicolumn{1}{l}{EPE FS ↓}  \\ 
\hline
2                               & 0.0528                        & 0.1222                     & 0.0055                     & 0.0308                      \\
4                               & \textbf{0.0516}               & 0.1212                     & 0.0047                     & 0.0289                      \\
8                               & 0.0517                        & 0.1214                     & 0.0042                     & 0.0295                      \\
16                              & 0.0532                        & 0.1262                     & 0.0038                     & 0.0297                      \\
\bottomrule
\end{tabular}
\caption{Ablation study on GRU iteration count. Analyzing the performance impact across varying GRU iteration numbers, with 4 iterations emerging as the optimal configuration for DeFlow.}
\label{tab:gru_itr}
\vspace{-1.0em} 
\end{table}
\begin{table*}[t!]
\centering
\def\arraystretch{1.2}
\begin{tabular}{ccccccccc} 
\toprule
Decoder Design                     & Res ($\mathrm{m}$) & EPE 3-Way ↓ & EPE FD ↓ & EPE BS ↓ & EPE FS ↓ & Dynamic IoU ↑ & GM ($\mathrm{MiB}$) ↓ & FPS  ↑  \\ 
\hline
\multirow{3}{*}{FastFlow3D~\cite{fastflow3d}} 
& 0.4 & 0.1116                        & 0.3055                     & 0.0037                     & 0.0254                     & 0.4701 & 2114 & 49.71                       \\
& 0.2 & 0.0852                        & 0.2326                     & 0.0025                     & 0.0206                     & 0.5257 & 2874 & 29.17                       \\
& \textcolor{red}{0.1} & \underline{0.0586}         & 0.1463      & 0.0086      & 0.0208      & 0.5238           & 
\textcolor{red}{6634} 
& \textcolor{red}{11.31}         \\
\cline{1-1}
Our w/o GRU            & 0.2 & 0.0916    & 0.2499  & 0.0034 & 0.0216 & 0.5224      & 2876 & 28.99   \\ 
\cline{1-1}
Ours                & 0.2 & \textbf{0.0564}               & 0.1309                     & 0.0045                     & 0.0337 & 0.4896      & 2878 & 20.49  \\
\bottomrule
\end{tabular}
\caption{Decoder design and resolution impact on Argoverse 2 sensor validation set. The table contrasts results based on different decoder designs and voxel resolutions. `Res' indicates voxel resolution, `GM' denotes GPU memory consumption during model execution, and `FPS' (frame per second) signifies the model's running speed. 
The findings reveal that the GRU at a 0.2$\mathrm{~m}$ resolution overperformance the original 0.1$\mathrm{~m}$ resolution decoder with reduced GPU memory requirements and faster running speed.}
\label{tab:resolution_model}
\end{table*}
\section{Experiment}
\label{sec:exp}
In this section, we outline our experimental setup, followed by a series of ablation studies to understand the contributions of our approach. We then present quantitative comparisons with state-of-the-art methods on a benchmark dataset and conclude with qualitative visualization results.

\subsection{Experiment Setup}
\label{sec:exp_set}
\textbf{Dataset}: Our approach is evaluated on the large-scale autonomous driving data, Argoverse 2~\cite{Argoverse2} which encompasses a variety of sets, including \textit{Sensor} and \textit{LiDAR}. Given that the \textit{LiDAR} dataset lacks imagery and any other annotations, our primary focus is on the \textit{Sensor} dataset.
The \textit{Sensor} dataset encompasses 700 training and 150 validation scenes. Each scene is approximately 15 seconds long in 10$\mathrm{~Hz}$, complete with annotations. Argoverse 2 provides an online benchmark with 150 testing scenes. 

\textbf{Methods}: 
Our main comparison is with FastFlow3D~\cite{fastflow3d} on the validation dataset. 
We reproduce this method based on Zeroflow~\cite{zeroflow}, given that FastFlow3D serves as its student network and its code is publicly available.
To showcase our method's superiority, we display results of various baseline methods from the online leaderboard, 
including NSFP~\cite{li2021neural}, FastNSF~\cite{li2023fast}, and Zeroflow~\cite{zeroflow}. NSFP, FastNSF operates on a self-supervised paradigm.
Both FastFlow3D and our Deflow employ supervised learning techniques, leveraging the \textit{Sensor} dataset.
Zeroflow adopts a semi-supervised strategy, utilizing NSFP for dataset labeling. While the standard version relies solely on the \textit{Sensor} dataset, the XL version incorporates additional data from the \textit{LiDAR} dataset, amounting to twice the original size. Furthermore, the XL version sets the resolution to $0.1\mathrm{~m}$ and boosts a model with ten times the parameters of the standard version.

\textbf{Implementation Details}: 
In~\cref{tab:main_res} for the Argoverse 2 test dataset, our DeFlow implementation is as follows: We use four GRU iterations (as shown~\cref{tab:gru_itr}), and the model trains for a total of 50 epochs with a batch size of 80. The network optimization leverages the Adam optimizer with a learning rate set at $2\times 10^{-6}$.  
The chosen loss function is detailed in~\cref{eq:scaled_three}. 
The resolution is set as $0.2\mathrm{~m}$, consquently, the $[512,512]$ grid is transformed from a $102.4\mathrm{~m}\times 102.4\mathrm{~m}$ map.
Other benchmarking methods in detail: FastNSFP uses their official public code with config and FastFlow3D trains in 50 epochs to converge and 64 batch size~\cite{fastflow3d} as same as their paper's setup. Others (ZeroFlow and NSFP) are public on the online leaderboard.
For local experiments, aimed at a fair ablation study, all models are trained with a learning rate of $2\times 10^{-6}$ and the same batch size of 80. We maintain strict control over the resolution variable to see the influence of resolution, loss function, and network design on performance. 
All local experiments were executed on a desktop powered by an Intel® Core™ i9-12900KF and equipped with a GeForce RTX 3090 GPU.

\textbf{Metric}: The Argoverse 2 benchmark adopts a unified metric proposed by~\cite{chodosh2023re}, which introduces the 3-way Endpoint Error (EPE). EPE, as defined in~\cref{eq:opti}, measures the L2 norm of the discrepancy between the predicted and actual flow vectors, expressed in meters. The 3-way calculates the unweighted average EPE across three classifications: Foreground Dynamic (FD), Foreground Static (FS), and Background Static (BS). 
If the flow of a point $\hat{\mathbf{F}} (p)$ exceeds $0.05\mathrm {~m}$, the point is defined as dynamic.
Given the dataset's 10 Hz collection frequency, this threshold corresponds to a speed of $0.5 \mathrm{~m} / \mathrm{s}$. The Dynamic Accuracy Relaxed (Dynamic AccRelax) metric captures the proportion of dynamic points with an EPE under $0.1\mathrm{~m}$ or a relative error below 10\%. In contrast, the Strict (Dynamic AccStrict) metric requires an EPE under $0.05\mathrm{~m}$ or a relative error of less than 5\%.

\subsection{Quantitative Results}
\label{sec:quan}
In this section, we evaluate the efficacy of our proposed Deflow approach and compare it with alternative methods, highlighting the effects of different design choices on performance.

As referenced in~\cref{tab:main_res} and briefly discussed in~\cref{sec:exp_set}, it's evident that our method achieves the state-of-art performance in the Argoverse 2 scene flow task.  The most significant enhancement is observed in the boosted accuracy of dynamic point flow estimation, with only minimal errors occurring in static points. 

\begin{figure*}[t!]
\centering
\includegraphics[trim=12 8 15 40, clip, width=\linewidth]{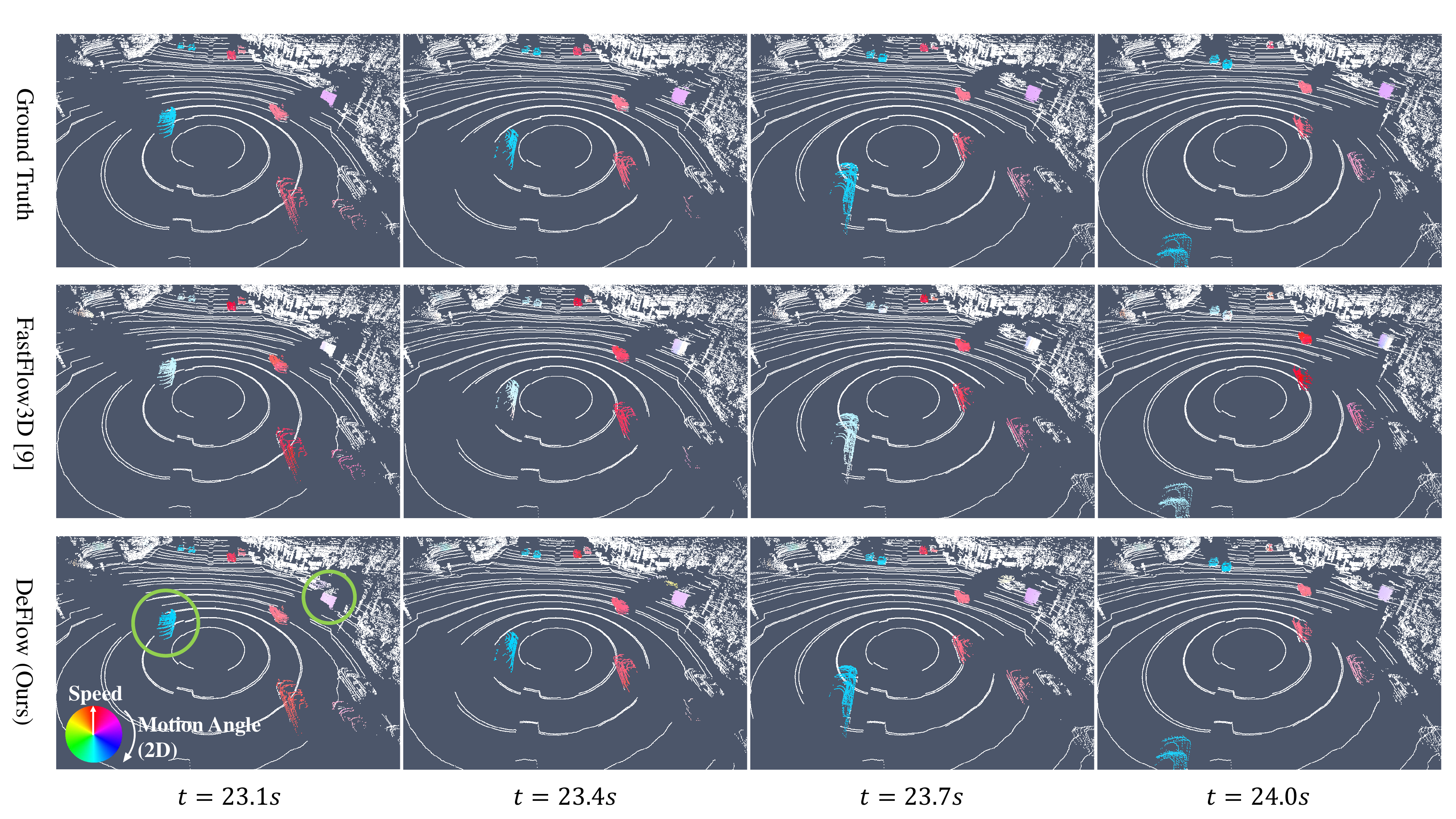}
\caption{Qualitative results from the validation dataset. The top row displays the ground truth flow, the middle row presents the FastFlow3D result, and the bottom row showcases the DeFlow outcomes. 
DeFlow estimates closely match the ground truth flow in both speed and angle. 
As highlighted in the two green circles, our DeFlow method demonstrates better performance in predicting motion angle (indicated by color variations) and speed (represented by color intensity) compared to FastFlow3D.
The color wheel has been adjusted to align with the ego vehicle's forward direction.}
\label{fig:q1}
\end{figure*}

\begin{table}[t]
\centering
\def\arraystretch{1.2}
\begin{tabular}{lcccc}
\toprule
Loss Type                           &EPE 3-way &EPE FD &EPE BS &EPE FS \\ 
\hline
FastFlow3D~\cref{eq:scaled_type} & 0.0852                        & 0.2326                     & 0.0025                     & 0.0206                      \\
Zeroflow~\cref{eq:scaled_speed}  & 0.0843                        & 0.2112                     & 0.0201                     & 0.0202                      \\
Ours~\cref{eq:scaled_three} & \textbf{0.0787}               & 0.2045                     & 0.0041                     & 0.0277    
\\
\bottomrule
\end{tabular}
\caption{Ablation study on loss types. Our proposed loss function outperforms others, achieving the best EPE 3-way score by addressing the imbalanced data distribution between static and dynamic points.}
\label{tab:loss}
\vspace{-1.0em} 
\end{table}

To illustrate the efficacy of our network design and the impact of different resolution settings, we evaluated a model without the GRU iteration module, as depicted in~\cref{fig:arc}. Additionally, we tested both smaller and larger resolutions in FastFlow3D. 
As shown in~\cref{tab:resolution_model}, where all models use the origin loss function~\cref{eq:scaled_type}, the results without the GRU module in the network indicate that merely extending the point offset feature and grid feature channels using MLPs can adversely affect the accuracy of dynamic point flow. 
Conversely, our Deflow GRU framework significantly reduces the EPE 3-way and enhances accuracy in both relaxed and strict dynamic metrics. 
When comparing with FastFlow3D in $0.1\mathrm{~m}$ resolution setting, with larger GPU memory consumption and slower FPS, our DeFlow GRU at $0.2\mathrm{~m}$ achieves better performance with efficiency on both GPU memory and computation speed.
Considering both performance and computational resources, our method is more suitable for lightweight on-board computation in real-time.
This outcome further highlights the effectiveness of our network design.

The design of the loss function plays an important role in training the network, as discussed in~\cref{sec:loss_fn}. 
We conducted an ablation study using the original FastFlow3D to assess various loss types. Apart from the loss functions, all other parameters were kept consistent: a training epoch of 50, a learning rate of $2\times 10^{-6}$ using the Adam optimizer, and a resolution of $0.2\mathrm{~m}$.
The loss proposed by us, as presented in~\cref{tab:loss}, emerged as the superior choice, reducing the EPE FD error by 12.1\% compared to FastFlow3D. 

\subsection{Qualitative Results} 
We showcase the qualitative outcomes of our DeFlow on the Argoverse 2 validation set with~\cref{fig:q1} serving as an example.
From our observation, DeFlow demonstrates the ability to accurately capture the motion flow in most cases. 
When compared to FastFlow3D, it performs better in the prediction of both speed and motion angle. 
In certain regions, particularly those that are blocked or partially occluded, DeFlow occasionally exhibits inaccuracies, indicating potential areas for further improvement.

\section{Conclusion}
In this paper, we introduce DeFlow, an efficient and high-performance method for autonomous driving in large-scale point clouds. Our primary contributions include the introduction of the DeFlow network, which enhances the extraction and reconstruction of point-voxel-point network features at the point level. Additionally, we propose a novel loss function to address the challenges of imbalanced data distribution among points. Our experimental results underscore the efficacy of our approach.

Future work could be on self-supervised exploration of DeFlow and the fusion with multi-modality sensors, like cameras and radar.
The flow of dynamic objects is one that we are mainly focused on in scene flow estimation, so there is a possible solution if we can segment static and dynamic~\cite{zhang2023benchmark} first, then it can hugely decrease the computation burden for neural optimization-based approaches.

\addtolength{\textheight}{-3cm}   
\section*{Acknowledgement}
Thanks to one of ZeroFlow authors: Kyle Vedder who kindly discussed their results with us and HKUST Ramlab's member: Jin Wu who gave constructive comments on this work. 
This work was partially supported by the Wallenberg AI, Autonomous Systems and Software Program (WASP) funded by the Knut and Alice Wallenberg Foundation and Prosense (2020-02963) funded by Vinnova. The computations were enabled by the supercomputing resource Berzelius provided by National Supercomputer Centre at Linköping University and the Knut and Alice Wallenberg Foundation, Sweden.

\bibliographystyle{IEEEtran}
\bibliography{IEEEabrv,ref}

\end{document}